\documentclass[letterpaper, 12pt]{article}
\usepackage{authblk}
\usepackage{times}
\usepackage[margin=1in]{geometry}
\usepackage[empty]{fullpage}
\usepackage{graphicx}
\usepackage{wrapfig}
\usepackage{caption}
\usepackage{url}
\usepackage{multirow, tabularx}
\usepackage{booktabs}

\usepackage[utf8]{inputenc}
\usepackage{footnote}
\makesavenoteenv{tabular}
\makesavenoteenv{table}
\usepackage{fancyhdr}
\usepackage{authblk}

\newcolumntype{L}[1]{>{\raggedright\arraybackslash}m{#1}}
\newcolumntype{C}[1]{>{\centering\arraybackslash}m{#1}}
\newcolumntype{R}[1]{>{\raggedleft\arraybackslash}m{#1}}

\title{A Comparative Analysis of Distributional Term Representations for Author Profiling in Social Media\footnote{Preprint   of   \cite{alvarez2019comparative}.
The   final   publication   is   available   at   IOS   Press   through https://content.iospress.com/articles/journal-of-intelligent-and-fuzzy-systems/ifs179033}}

\author[1]{Miguel Á. Álvarez-Carmona}
\author[2]{Esaú Villatoro-Tello}
\author[1]{Manuel Montes-y-Gómez}
\author[1]{Luis Villaseñor-Pienda}
\affil[1]{Computational Sciences Department, Instituto Nacional de Astrof\'{i}sica, \'{O}ptica y Electr\'{o}nica, Luis Enrique Erro 1, Puebla 72840, M{\'e}xico}
\affil[2]{Information Technologies Department, Universidad Aut\'onoma Metropolitana, Unidad Cuajimalpa (UAM-C), Ciudad de M\'exico, 05348, M\'exico}

\begin{document}

\maketitle
\pagestyle{fancy}
\begin{abstract}
Author Profiling (AP) aims at predicting specific characteristics from a group of authors by analyzing their written documents. Many research has been focused on determining suitable features for modeling writing patterns from authors. Reported results indicate that content-based features continue to be the most relevant and discriminant features for solving this task. Thus, in this paper, we present a thorough analysis regarding the appropriateness of different distributional term representations (DTR) for the AP task. In this regard, we introduce a novel framework for supervised AP using these representations and, supported on it. We approach a comparative analysis of representations such as DOR, TCOR, SSR, and word2vec in the AP problem. We also compare the performance of the DTRs against classic approaches including popular topic-based methods. The obtained results indicate that DTRs are suitable for solving the AP task in social media domains as they achieve competitive results while providing meaningful interpretability.

\end{abstract}


\section{Introduction}
\label{sec:Intro}
The great amount of social media content shared on the Internet has motivated the development of new methods and tools for its analysis and usage. In particular, profiling social media users is becoming an important issue for many companies and organizations. For example, from the marketing perspective, knowing characteristics of a group of Internet users could help in improving the impact of some particular products, and, from the forensic linguistics view, knowing the linguistic profile of an author could be used as valuable additional evidence in criminal investigations.


Generally speaking, the author profiling (AP) task consists in analyzing written documents to extract relevant demographic information from their authors \cite{koppel2002automatically}, such as gender, age range, personality traits, native language, political orientation, among others. 

Traditionally, the AP task has been approached as a single-labeled classification problem, where the different categories (e.g., \textit{male} vs. \textit{female}, or \textit{teenager} vs. \textit{young} vs. \textit{old}) stand for the target classes. The common pipeline is as follows: \textit{i}) extracting textual features from the documents; \textit{ii}) building the documents' representation using the extracted features, and \textit{iii}) learning a classification model from the built representations. As it is possible to imagine, extracting the relevant features is a key aspect for learning the textual patterns of the different profiles. Accordingly, previous research has evaluated the importance of thematic (content-based) features \cite{koppel2002automatically,poulston13using} and stylistic characteristics \cite{bergsma2012stylometric}. 
More recently, some works have also considered learning such representations utilizing Convolutional and Recurrent Neural Networks  \cite{sierra2018combining,kodiyan2017author,takahashi2018text}. 

Although many textual features have been used and proposed, a common conclusion among previous research is that content-based features are the most relevant for this task. The later can be confirmed by reviewing the results from the PAN\footnote{A set of shared tasks on digital text forensics: {http://pan.webis.de/}} competitions \cite{rangel2018overview}, 
where the best-performing systems employed content-based features for representing the documents regardless of their genre. This result is somehow intuitive since AP is not focused on distinguishing a particular author through modeling its writing style\footnote{A problem known as Authorship Attribution.}, but on characterizing a group of authors. For example, in \cite{schler2006effects} authors performed an exhaustive study of non-formal documents in order to determine the pertinence of content-based features. They found that stylistic features do not provide any additional information to the learning algorithm. In contrast, content words such as \textit{linux} and \textit{office}, and \textit{love} and \textit{shopping}, showed to be highly discriminant for males and females respectively. 
%
%


In line with these findings, our previous research has focused on evaluating the pertinence of distinct content-based representations for solving the AP task. Mainly, we went beyond the traditional bag of words by considering distributional and topic-based representations. The idea behind both approaches was to develop enriched representations that help to overcome the small-length and high-sparsity issues of social media documents by considering contextual information computed from document occurrence and term co-occurrence statistics. Mainly, we proposed a family of distributional representations based on second order attributes which allow capturing the relationships between terms and profiles and subprofiles \cite{LopezMonroy2015134}. These representations obtained the best results in the AP tasks at PAN 2013 and PAN 2014 \cite{pastor14}. Also, we evaluated topic-based representations such as Latent Semantic Analysis (LSA) and Latent Dirichlet Allocation (LDA) in the AP task \cite{alvarezinaoe}, obtaining the best performance at the PAN 2015 as well as showing its superiority against a representation based on manually defined topics utilizing LIWC \cite{alvarez2016evaluating}.

Motivated by the good results of our subprofile-specific representation (SSR) \cite{LopezMonroy2015134}, as well as by the recent use of word embeddings in the AP task \cite{lopez2018custom}, in this paper we present a thorough analysis on the pertinence of \textit{distributional term representations} (DTRs) for solving the problem of AP in social media. We aim to highlight the advantages and disadvantages of this type of representations in comparison with traditional topic-based representations such as LSA and LDA.

%
%
%
In summary, the main contributions of this paper are:
\begin{itemize}
\item We introduce a framework for supervised author profiling in social media domains using DTRs. This framework encompasses the extraction of distributional representation of terms as well as the construction of the authors' representation by aggregating the representations of the terms from their documents.
\item We evaluate for the first time the document-occurrence representation (DOR) and the term co-occurrence representation (TCOR) in the AP task. These are two simple and well-known term representations from distributional semantics \cite{Lavelli:2004:DTR:1031171.1031284}. 
\item We present a comparative analysis of several distributional representations, namely DOR, TCOR, SSR, and word2vec, using the proposed framework for AP. Additionally, we compare their performance against the results from classic bag-of-words and topic-based representations. 
\end{itemize}
%
%
For evaluating the proposed framework and performing the analysis of the distinct DTRs, we employed the PAN@2014 dataset. This corpus was specially built for studying AP in social media domains, as it contains data from blogs, Twitter, and reviews. 
We performed several experiments aiming at determining the suitability of DTRs for solving the AP task in social media domains. Our initial intuitions suggest that through the use of these representations will be possible to obtain richer content-based features as well as --for some of them-- easily interpretable results. Thus, we carry out an analysis of the obtained results and their relation to different characteristics of the considered text collections such as their lexical complexity, shortness, and class imbalance.


The rest of the paper is organized as follows: Section \ref{sec:RelatedWork} describes some previous work in AP. Section \ref{sec:ProposedFramework} explains the proposed framework for AP in social media. It also describes the distributional and distributed term representations employed in our study. Section \ref{sec:Dataset} provides a brief description of the selected dataset. Section \ref{sec:Results} explains the experimental settings and shows the evaluation results. Finally, Section \ref{sec:Conclusions} presents our conclusions and depicts some future work directions.

\section{Related work}
\label{sec:RelatedWork}
As previously mentioned, the AP task consists in analyzing texts to predict general or demographic attributes of authors such as their gender, age, personality, native language, political orientation, among others.

The AP has been a relevant research topic for several years now. As an example of this, we can refer to the PAN series celebrated from 2012 to 2018 \cite{rangel2018overview}, where distinct AP methods have been tested under strict conditions. Results from these tracks show that the AP task is still an unsolved problem, and there is enough room for improvement.

Although there is a wide range of methods for approaching the AP task, traditionally it has been tackled as a text classification problem \cite{sebastiani}. Under this scenario, the primary focus of research has been on the selection of the best features for modeling the authors' profiles. Mainly, two types of textual features have played a key role: \textit{i}) content-based features, such as word n-grams and topic-based models, and \textit{ii}) style-based features such as function words, punctuation marks, and emoticons. Examples of the former approach are the classic work \cite{koppel2002automatically}, where a bag of words representation was used for gender classification; and the recent work \cite{basile2017n}, where word n-grams were used as primary features.
On the other hand, the works \cite{argamon2005lexical} and \cite{kiprov:2015} are two relevant examples of works evaluating a broad family of stylistic features for determining age, gender and personality traits of social media users.
In general, the results from the PAN evaluation forums indicate that most successful works for AP in social media have used combinations of these two kinds of features. However, there is a clear tendency to use style-based features just as a complement to the content-based features.

Moving a step forward from the traditional bag of words representation, some works have used topic-based representations in combination with other different features for capturing more sophisticated content-based features. For instance, \cite{meina2013ensemble} evaluated several structural and style-based features in combination with LSA. Similarly, \cite{schwartz2013personality} combined the features extracted using LDA with those from the lexical resource LIWC. In \cite{alvarezinaoe} second order attributes were used in combination with LSA for age and gender identification. \cite{poulston2015topic} employed POS, word n-grams, and LDA features, indicating that the combination of word n-grams with LDA yields better results. \cite{farseev2015harvesting} combined behavioral, location, as well as textual features extracted with LDA.
Contrary to \cite{poulston2015topic}, it concluded that the LDA results are not that useful. \cite{mccollister2016predicting} is another work that concluded that LDA is not a suitable method for AP. Both LSA and LDA have also been evaluated under a cross-genre scenario \cite{markov2016adapting,bilan2016caps}, however, there is not a clear conclusion regarding the importance of topic-modeling under these circumstances. In summary, there is no clear evidence about the importance of the topic-based methods in solving the posed task, since most of the reported results are done in combination with different strategies, and the lack of an individual analysis avoids to reach solid conclusions. 

In a different research direction, other works have proposed using distributional term representations for modeling more discriminative content-based features. For example, \cite{LopezMonroy2015134} employed second order attributes for capturing discriminative sub-profile specific information of terms. Some other approaches have incorporated the distributional information through the use of word embeddings, namely word2vec and doc2vec. For example, \cite{franco2015language} proposed a methodology for language variety identification, and \cite{bayot2016author} employed an SVM learning algorithm in combination with a doc2vec representation for AP. In a similar direction, the works described in \cite{poulston13using,gomez2016improving} proposed different ways for computing the word embeddings and then used them as features for training a supervised classifier. More recently, some works have used different deep learning models and strategies for learning representations for AP. For example, Neural Attention Models \cite{miura13author}, Recurrent Neural Networks \cite{kodiyan2017author,takahashi2018text}, Subwords embeddings \cite{franco2017subword}, and Convolutional Neural Networks \cite{sierra2018combining}. 
In general, these works have confirmed that content features (e.g., words and word n-grams) perform better than style features (characters, char n-grams, etc.).

%
%

In summary, from the review of the current literature of AP we could identify the following issues: \textit{i}) several recent works indicate that content-based features are more relevant than style-based features; \textit{ii}) sophisticated deep-learning strategies have not been able to consistently and significantly outperform traditional approaches
; \textit{iii}) there is no a thorough analysis regarding the pertinence of topic-based methods for solving the AP task and, more importantly, \textit{iv}) some distributional term representations have been recently used in AP showing good results; nevertheless, some classic representations such as DOR and TCOR have not been evaluated yet. Hence, this paper aims to thoroughly examine the suitability of DTRs for AP in social media domains and their competence in comparison to topic-based representations.

\section{Distributional framework for AP}
\label{sec:ProposedFramework}

This Section describes a general framework for Author Profiling using distributional term representations (DTRs). By exploiting DTRs, we aim to represent documents\footnote{Hereafter we are going to use the term \textit{document} as a synonym of \textit{user}, under the consideration that all the posts from a user form a document.} from social media users in a non-sparse space, which captures more discriminative information. Our goal is to overcome, to some extent, the issues naturally inherited by the BoW and build instead a more semantically related representation. Intuitively, DTRs can capture the semantics of a term $t_i$ by exploiting the distributional hypothesis; ``words with similar meanings appear in similar contexts''. Thus, different DTRs can capture the semantics through the context in different ways and at different levels.

\begin{figure}
  \centering
    \includegraphics[width=5in]{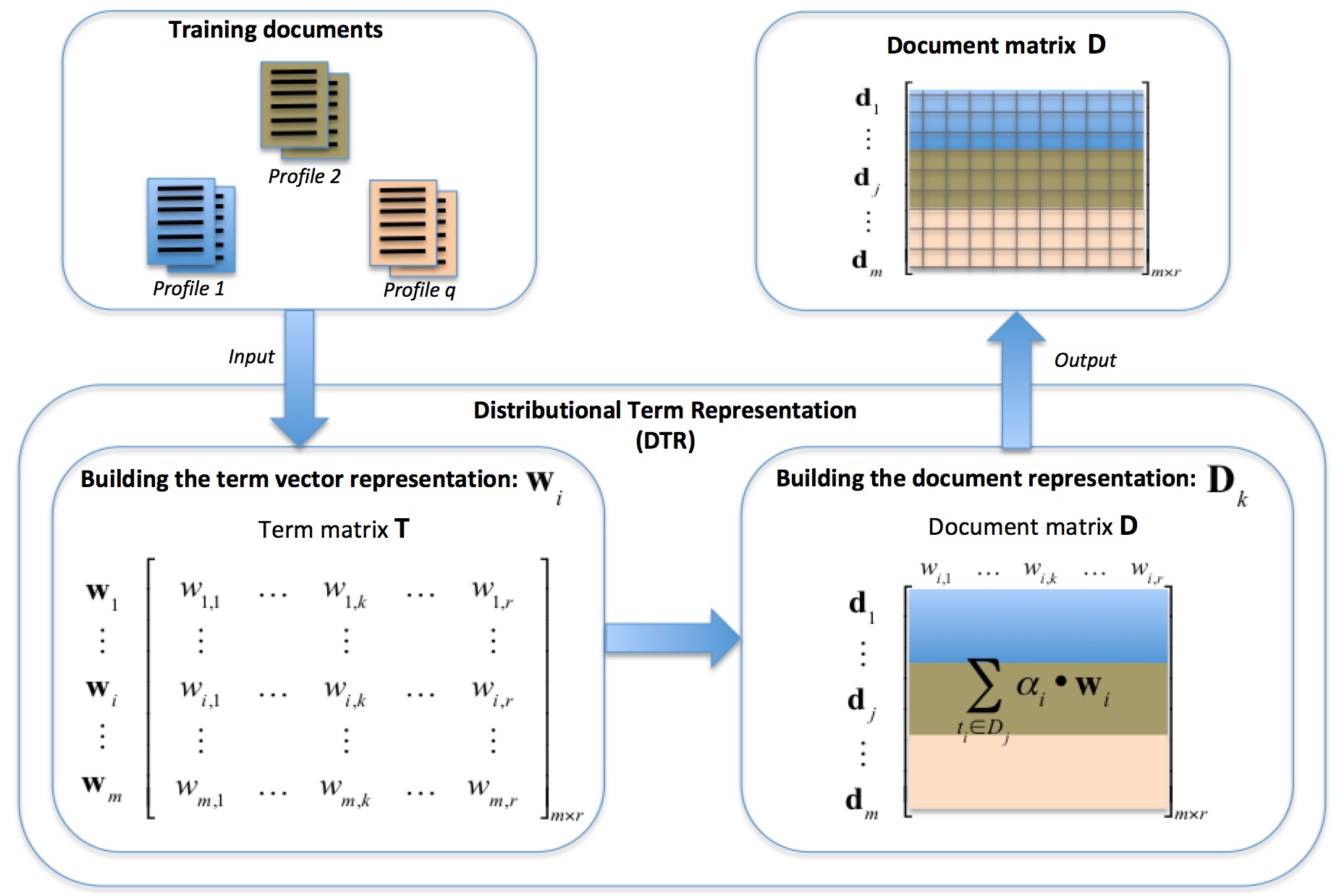}
    \caption{General diagram of the proposed framework for building the Distributional term representation of documents.}
  \label{fig:GeneralFramework}
\end{figure}

The proposed framework is shown in Figure \ref{fig:GeneralFramework} and it comprises two main stages: \textit{i}) to determine the terms' vector representations, and \textit{ii}) to build the document representations. Notice that term representations account for discriminative semantic relationships between terms. Then, document representations are obtained by aggregating the representations of terms that occur in each document, leading to a distributional-based representation. The resultant document representations are non-sparse and capture useful profile information. The way in which terms and documents are represented in each stage is the same. The only difference is on how the semantics of each term is determined, i.e., how the DTR is computed. Once documents are represented, a standard classifier is considered to build an AP model. The rest of the section details the way in which terms and documents are represented, and how the distinct DTRs are obtained.

\subsection{Distributional term representations}
As mentioned, our proposed framework to represent a document using DTRs requires of two steps (Figure \ref{fig:GeneralFramework}). The first step consists in building a vector representation for each term. The second step uses these term vectors to generate the document representation. To put things simple, lets consider words in the vocabulary as the base terms for building the DTR. More formally, let $\mathcal{D}=\{(d_1, y_1),\ldots, (d_n,y_n)\}$ be a training set of $n-$pairs of documents ($d_j$) and labels/categories $y_i \in \mathcal{C} = \{C_1,\ldots,C_q\}$. Also let $\mathcal{V} = \{t_1,\ldots,t_m\}$ be the collection vocabulary. In this context, DTRs associates each term $t_i \in \mathcal{V}$ with a term vector $\vec{w_i} \in {R}^r$, i.e., $\vec{w_i}= \langle w_{i,1},\ldots, w_{i,r}\rangle$. In this notation $w_{i,j}$ indicates the contribution of distributional feature $j$ to the representation of term $t_i$. This contribution is particular of each DTR and can be computed in a number of ways. In the following sections we describe in detail each of the DTRs that we selected for this study. The second step consists in building the document representations by using the term vectors. More formally, the representation of document $d_j$, the vector $\vec{d_j}$, is obtained by using Equation \ref{eq:dtrsdoc}, where the scalar $\alpha_{i}$ weights the relevance of term $t_i$ in document $d_j$. Although there are a number of ways to define this weighting, the most widely used approach is the average (i.e., $\alpha_{i}$ is proportional to the number of terms in the document).

\begin{equation}
\label{eq:dtrsdoc}
 \vec{d_{j}}=\sum_{t_i \in d_j} \alpha_{i} \cdot \mathbf{w_i} 
\end{equation}

\subsubsection{Document occurrence representation}
\label{subsec:dor}
The document occurrence representation (DOR) can be considered the dual of the TF-IDF representation widely used in the Information Retrieval field \cite{lavelli2004distributional}. DOR is based on the hypothesis that the semantics of a term can be revealed by its distribution of occurrence-statistics over the documents in the corpus. A term $t_i$ that belongs to the vocabulary $\mathcal{V}$ is represented by a vector of weights associated to documents $\vec{w_i}
= \left \langle w_{i,1}, \cdots ,  w_{i,N} \right \rangle$
where $N$ is the number of documents in the collection and $0 \le w_{i,j} \le 1$ represents the contribution of document $d_j$ to the specification of the semantics of $t_i$:

\begin{equation}
w_{i,j} = df(t_i,d_j)\log{\frac{|\mathcal{V}|}{N_j}}
\end{equation}
where $N_j$ is  the  number  of  different  terms  from  the  dictionary $\mathcal{V}$ that  appear  in document $d_j$, $|\mathcal{V}|$ is the number of terms in the vocabulary, and 
\begin{equation}
df(t_i,d_j) = \left\{
        \begin{tabular}{cc}
            $1+\log(\#(t_i,d_j))$ & if $\#(t_i,d_j) \ge 0$ \\
            $0$ & if $otherwise$ \\
        \end{tabular}
\right.
\end{equation}
where $\#(t_i,d_j)$ denotes the number of times term $t_i$ occurs in document $d_j$. Intuitively, the more frequent the term $t_i$  is in document $d_j$, the more important is $d_j$ to characterize the semantics of $t_i$. In the same way, the more terms contain $d_j$, the less its contribution in characterizing the semantics of $t_i$.

\subsubsection{Term co-occurrence representation}
\label{subsec:tcor}
Term Co-Occurrence Representation (TCOR) is based on co-occurrence statistics \cite{lavelli2004distributional}. The underlying idea is that the semantics of a term $t_i$ can be revealed by the terms that co-occur with it across the documents collection. Here, each term $t_i \in \mathcal{V}$  is represented by a vector of weights $\vec{w_i} = \left \langle w_{i,1}, \cdots ,  w_{i,|\mathcal{V}|} \right \rangle$ where  $0 \le w_{i,j} \le 1$ represents the contribution
of term $t_j$ to the semantic description of $t_i$, and is computed as follows:

\begin{equation}
w_{i,j} = tf(t_i,t_j)\log{\frac{|\mathcal{V}|}{\mathcal{V}_k}}
\end{equation}
where $\mathcal{V}_k$ is the number of different terms in the dictionary $\mathcal{V}$ that co-occur with $t_i$ in at least one document, and:
\begin{equation}
tf(t_i,t_j) = \left\{ 
        \begin{tabular}{cc}
            $1+\log(\#(t_i,t_j))$ & if $\#(t_i,t_j) \ge 0$ \\
            $0$ & if $otherwise$ \\
        \end{tabular}
\right.
\end{equation}
where $\#(t_i,t_j))$ denotes the number of documents in which term $t_j$ co-occurs with the term $t_i$.

\subsubsection{Word embeddings: word2vec}
\label{subsec:doc2vec}
Recently, a very popular group of related models for producing word embeddings is the word2vec~\cite{mikolov2013distributed}. These models are shallow, two-layer neural networks trained to reconstruct the linguistic contexts of words. Word2vec takes as its input a large corpus of texts and produces a vector space, typically of a few hundred dimensions, where each term in the corpus is assigned to a corresponding vector $\vec{w_i}$ in the space. Thus, once the word vectors have been computed and positioned in the vector space, words that share common contexts in the corpus are located close to one another in the space \cite{mikolov2013efficient}.

Word2vec employs either one of two model architectures to produce the distributed representation of words: \textit{i}) continuous bag-of-words (CBOW), or \textit{ii}) continuous skip-gram \cite{mikolov2013efficient}. In the continuous bag-of-words architecture, the model predicts the current word from a window of surrounding context words. The order of context words does not influence prediction. In the continuous skip-gram architecture, the model uses the current word to predict the surrounding window of context words. The skip-gram architecture assigns a higher weight to those nearby context words while more distant context words are considered less important \cite{mikolov2013distributed}. The CBOW model is faster than the skip-gram model. However, the later does a better job for handling infrequent words \cite{zhuang2017challenges}. 

In our experiments we built the word embeddings (i.e., vectors $\vec{w_i}$) using the skip-gram model. It mainly considers the conditional probabilities $p(c|t)$ for all terms $t$ and their respective contexts $c$. Thus, given a corpus $D$, it aims to set the parameters $\theta$ of $p(c|t;\theta)$ so as to maximize the corpus probability:

\begin{equation}
argmax_\theta \prod_{(t,c)\in D} p(c|t;\theta)
\end{equation}
%
%


In the end, the purpose of  word2vec is to build an accurate representation of words in a space ${R}^r$, where similar vectors correspond to semantically related words \cite{mikolov2013efficient}. For example, the \textit{Paris} vector must be close to the \textit{Berlin} vector, since both are capitals, so as the vectors from \textit{elephant} and \textit{dog} since both are animals.


\subsubsection{Subprofile specific representation}
\label{subsec:ssr}

The intuitive idea of the second order attributes consists in representing the terms by their relation with each target class \cite{li2011fast,LopezMonroy2015134}. This can be done by exploiting occurrence-statistics over the set of documents in each one of the target classes. In this way, we represent each term $t_i \in \mathcal{V}$ with a vector $\vec{w_i} = \left\langle w_{i,1} , \cdots , w_{i,q} \right\rangle$, where the scalar $w_{i,k}$ is the degree of association between word $t_i$ and class $C_k$. Under this DTR, the weight $w_{i,k}$ is directly related to the number of occurrences of term $t_i$ in documents that are labeled with class $C_k$. The relationship between the $i^{th}$ word and the $k^{th}$ class can be defined according to:

\begin{equation}
\label{eq:SOA}
w_{i,k} = \sum_{\forall d_j:y_j == C_k} \log_{2} \left (1 + \frac{tf(t_i,d_j)}{len(d_j)} \right )
\end{equation}

where $tf(t_i, d_j)$ is the occurrence frequency of the word $t_i$ in the document $d_j$, and $len(d_j)$ indicates the number of words in $d_j$. The $\log_2$ function aims to soften the relevance of highly frequent words. In our case, the classes are the different profiles that we aim to identify. Thus, $d_j$ represents documents that were produced by users with the same profile, e.g., same gender or same age rate.

The computed raw weights $w_{i,k}$ from Equation \ref{eq:SOA} can be directly used to build the term vectors. However, a term representation based on raw $w_{i,k}$ weights is sensitive to highly unbalanced data. Thus, in order to produce the final $\vec{w_i}$ representation, we consider applying two normalizations: a kind of row-based normalization to consider the proportion of the $|V|$ terms in each class, and then a kind of column-based normalization to take into account the weights computed for the $|C|$ classes, making weights $w_{i,k}$, comparable among classes. In the end, $\vec{w_i}$ can be seen as a probability distribution of $t_i$ over the distinct $k$ author profiles.

In \cite{LopezMonroy2015134}, second order attributes were modeled at sub-profile level; mainly, it was proposed to cluster the instances from each target in order to generate several subclasses. The idea was to consider the high heterogeneity of social media users. Utilizing this process, the set of target classes $C$ will now correspond to the set of all subgroups from the original target classes. This new representation is called \textit{Subprofile-based Representation} (SSR), and is considered one of the state-of-the-art representations for AP.




\section{Dataset description}
\label{sec:Dataset}
For our experiments we employed the English dataset from the PAN@2014 AP track\footnote{{http://pan.webis.de/clef14/pan14-web/author-profiling.html}}. This corpus was specially built for studying AP in social media. It is labeled by gender (i.e., female and male), and five non-overlapping age categories (18-24, 25-34, 35-49, 50-64, 65+). Although all documents are from social media domains, four distinct genres were provided: blogs, social media, hotel reviews, and Twitter posts. A more detailed description of how these datasets were collected can be found in \cite{rangel2014overview}.
Table \ref{tab:DataDistribution} provides some basic statistics regarding the distribution of profiles across the different domains (i.e., genres). It can be noticed that gender classes are balanced, whereas for the age classification task the classes are highly unbalanced. Particularly there are very few instances for the 65+ class.

\begin{table}[h!]
\caption{Distribution of the gender and age classes across the different social media domains.}
\centering
\begin{tabular}{lcccc}
\toprule 
\multirow{2}{*}{\textbf{Classes}} & \multicolumn{4}{c}{\textbf{Genres}}\\
\cmidrule(r){2-5}
 & \textit{Blogs} & \textit{Reviews} & \textit{Social-media} & \textit{Twitter} \\ 
\midrule
Female& 73& 2080& 3873 &153\\
Male & 74 &2080&3873 & 153 \\ 
\midrule 
\textit{Total:} &147 &4160  & 7746 & 306\\ 
\midrule
\midrule
18-24&6& 360& 1550  &20\\
25-34 &60 & 1000&2098 & 88 \\ 
35-49 &54& 1000& 2246&130\\
50-64  &23& 1000& 1838&60\\
65+ &4& 800& 14&8\\
\midrule
\textit{Total:} &147 &4160  & 7746 & 306\\ 
\bottomrule 
\end{tabular}
\label{tab:DataDistribution}
\end{table}




\section{Experiments and results}
\label{sec:Results}
This section explains the experiments that were carried out using the proposed framework. As we have previously mentioned, we aim at determining the pertinence of distributional term representations (DTRs) to the AP task in distinct social media domains. Accordingly, this section is organized as follows: firstly, Subsection \ref{subsec:setup} explains the experimental settings for all the experiments, then, Subsection \ref{subsec:results} describes the results obtained by each DTR in the four different social media domains.

\subsection{Experimental setup}
\label{subsec:setup}
%
%
\noindent\textbf{Preprocessing}: For computing the DTRs of each social media domain we considered the 10,000 most frequent terms. We did not remove any term, i.e., we preserved all content words, stop words, emoticons, punctuations marks, etc. In one previous work \cite{LopezMonroy2015134}, we demonstrated that preserving only the 10,000 most frequent words is enough for achieving a good representation of the documents.

\noindent\textbf{Text representation}: The different DTRs were computed as described in Section \ref{sec:ProposedFramework}. For the particular case of the word2vec representation, we employed two distinct configurations: \textit{w2v-wiki}, where the model was trained using a Wikipedia dataset, and \textit{w2v-sm}, where we trained a model for each one of the domains using the available training documents. In both cases, we used the word2vec skip-gram architecture as suggested in \cite{zhuang2017challenges}. Regarding the representation of the documents, in all cases, for all DTRs, we built their vectors by averaging the vectors from their words.

\noindent\textbf{Classification}: Following the same configuration than in our previous works (please refer to \cite{LopezMonroy2015134}), in all the experiments we used the linear Support Vector Machine (SVM) from the LIBLINEAR library with default parameters. 

%
%
%

\noindent\textbf{Baseline}: As baseline we employed the traditional bag-of-words (BoW) representation. 
We also compared the results from the different DTRs to those obtained by topic-based representations such as LSA and LDA as well as to those from the top systems from the PAN@2014 AP track.

%
%

\noindent\textbf{Evaluation}: We performed a stratified 10 cross-fold validation (10-CFV) strategy. For comparison purposes, and following the PAN guidelines, we employed the accuracy as the main evaluation measure.
Finally, we evaluated the statistical significance of the obtained results using a 0.05 significance level utilizing the Wilcoxon Signed-Ranks test.

\subsection{Results}
\label{subsec:results}
This section is organized as follows: firstly, we show the results from different DTRs for the age and gender classification tasks; secondly, we compare them against some topic-based representations and the best approaches from PAN 2014.

%
%

\subsubsection{Age and gender identification using DTRs}

Table \ref{tab:AccuracyAgeAllResults} and Table \ref{tab:AccuracyGenderAllResults} show the obtained results for the \textit{age} and \textit{gender} classification problems respectively. Each row represents one of the described DTRs, i.e., DOR, TCOR, word2vec, and SSR, while the last row represents the baseline results. Every column refers to a distinct social media genre, and the last column (i.e., \textit{Average}) represents the average performance for each method across all genres. In these tables, the best results are highlighted using boldface, and the star symbol ($\star$) indicates the differences that are statistically significant concerning the baseline results (in accordance to the used test; for details refer to Section \ref{subsec:setup}).


\begin{table}
 \centering
 \caption{Accuracy results obtained by the DTRs for the \textit{age} classification problem. Last column depicts the average performance of each approach across the distinct genres.}
 \begin{tabular}{lccccc}
 \toprule
 \multirow{2}{*}{\textbf{App.}} & \multicolumn{4}{c}{\textbf{Genres}}& \multirow{2}{*}{\textit{\textbf{Average}}}\\
 \cmidrule(r){2-5}
  & Blogs & Reviews & Social Media & Twitter& \\ 
 \midrule
 DOR &\textbf{0.49}$^\star$ & \textbf{0.36}$^\star$ & \textbf{0.38}$^\star$&0.47$^\star$&\textbf{0.42}\\
 TCOR & 0.38$^\star$&0.31 &0.32 &0.35&0.34\\
 w2v-wiki & 0.37&0.31 &0.36$^\star$ &0.43&0.36\\
 w2v-sm & 0.38$^\star$&0.30 &0.36$^\star$ &0.41&0.36\\
 SSR &0.48$^\star$  &0.34$^\star$  & 0.37$^\star$& \textbf{0.48}$^\star$&0.41\\ 
 \midrule
 \textit{Baseline}&0.34 & 0.28&0.32 &0.42&0.34\\
 \bottomrule
 \end{tabular}
 \label{tab:AccuracyAgeAllResults}
 \end{table}

Obtained results indicate that all DTRs, except for TCOR, outperformed the baseline method. In particular, DOR and SSR show statistically significant differences. These two methods obtained comparable results, being DOR slightly better than SSR in 5 out of 8 classification problems, which is an interesting result since SSR was among the winning approaches at PAN 2014. On the other hand, we attribute the low accuracy results showed by TCOR to the strong expansion that it imposes to the document representations. Considering direct term co-occurrences causes the inclusion of many unrelated and unimportant terms in the document vectors, and, therefore, difficulties the extraction of meaningful and straightforward profiling patterns.



 \begin{table}
 \centering
 \caption{Accuracy results obtained by the employed DTRs for the \textit{gender} classification task. Last column depicts the average performance of each approach across the distinct genres.}
 \begin{tabular}{lccccc}
 \toprule
 \multirow{2}{*}{\textbf{App.}} & \multicolumn{4}{c}{\textbf{Genres}}& \multirow{2}{*}{\textit{\textbf{Average}}}\\
 \cmidrule(r){2-5}
  & Blogs & Reviews & Social Media & Twitter& \\ 
 \midrule
 DOR & \textbf{0.78}$^\star$&\textbf{0.69}$^\star$ &0.52 &0.70&0.66\\
 TCOR &0.56 &0.62 &0.41 &0.54&0.53\\
 w2v-wiki & 0.75$^\star$&0.64 &0.52 &0.69&0.65\\
 w2v-sm & 0.74&0.64 &0.54&0.66&0.64\\
 SSR &\textbf{0.78}$^\star$  &\textbf{0.69}$^\star$  & \textbf{0.55}$^\star$& \textbf{0.71}&\textbf{0.68}\\ 
 \midrule
 \textit{Baseline}&0.72 & 0.62&0.52 &0.70&0.64\\
 \bottomrule
 \end{tabular}
 \label{tab:AccuracyGenderAllResults}
 \end{table}

Finally, another essential aspect to notice is the fact that both \textit{w2v-wiki} and \textit{w2v-sm} obtained similar results in each of the classifications problems, although the former learned the embeddings from a corpus that is not thematically and neither stylistically similar to the social media content. We presume these results could be explained by the relatively small size of the social media training collections, and, at the same time, by the large size and broad coverage of the used Wikipedia dataset, which has a vocabulary of 1,033,013 words.



\subsubsection{DTRs vs. topic-based representations}

Tables \ref{tab:AccuracyAgeVsTopicBased} and \ref{tab:AccuracyGenderVsTopicBased} compare the results from DOR and SSR, the best DTRs according to the previous results, against the results from two well-known topic-based representations, namely LDA and LSA. For both topic-based representations, the tables only show their the best result in each domain obtained after evaluating a different number of topics. The results marked with a $\ddagger$ indicate that they are significantly better than LSA, whereas results marked with $\dagger$ indicate that they are significantly better than LDA. Details on the test of statistical significance are given in Section \ref{subsec:setup}.


The obtained results clearly show that LDA was unable to obtain good results in both classification problems. This performance is in line with our previous findings reported in \cite{alvarez2016evaluating}. We hypothesize this poor performance is due to the dataset sizes;  bigger corpora are needed for extracting relevant and discriminative topics. 

\begin{table}
 \centering
 \caption{Comparison of the best DTRs against topic-based methods in the \textit{age} classification task. The last column shows the average performance of each approach across the different genres.}
 \begin{tabular}{lccccc}
 \toprule
 \multirow{2}{*}{\textbf{App.}} & \multicolumn{4}{c}{\textbf{Genres}}& \multirow{2}{*}{\textit{\textbf{Average}}}\\
 \cmidrule(r){2-5}
  & Blogs & Reviews & Social Media & Twitter& \\ 
 \midrule
  DOR &\textbf{0.49}$^\dagger$ & 0.36$^\dagger$ & \textbf{0.38}$^\dagger$$^\ddagger$&0.47$^\ddagger$&\textbf{0.42}\\
 SSR &0.48$^\dagger$  &0.34$^\dagger$  & 0.37$^\ddagger$& \textbf{0.48}$^\dagger$$^\ddagger$&0.41\\ 
 \midrule
 LDA & 0.44&0.27 & 0.37&0.47&0.38\\
 LSA & \textbf{0.49}&\textbf{0.37} &0.36 &0.45&\textbf{0.42}\\
  \midrule
  \midrule
 \cite{maharjan2014simple}&0.38&0.33&0.36&0.44&0.37\\
 \cite{villena2014daedalus}&0.39&0.31&0.35&0.41&0.36\\
 \cite{weren2014examining}&0.45&0.37&0.42&0.52&0.44\\
 \bottomrule
 \end{tabular}
 \label{tab:AccuracyAgeVsTopicBased}
 \end{table}
 
Regarding the LSA results, it is possible to observe, on the one hand, that for \textit{age} classification (refer to Table \ref{tab:AccuracyAgeVsTopicBased}), its average performance is similar to the one from DOR, i.e., 42\%. However, the only domain in which LSA outperforms DOR is in the reviews dataset. Nonetheless, there is no significant difference between these results. On the other hand, for \textit{gender} classification (Table \ref{tab:AccuracyGenderVsTopicBased}), LSA was not able to improve any result from DOR and SSR. It is important to mention that, although their results are comparable, LSA is a parametric method, and, therefore, tunning is required. 


 \begin{table}
 \centering
 \caption{Comparison of best DTRs against topic-based methods in the \textit{gender} classification task. The last column depicts the average performance of each approach across the different genres.}
 \begin{tabular}{lccccc}
 \toprule
 \multirow{2}{*}{\textbf{App.}} & \multicolumn{4}{c}{\textbf{Genres}}& \multirow{2}{*}{\textit{\textbf{Average}}}\\
 \cmidrule(r){2-5}
  & Blogs & Reviews & Social Media & Twitter& \\ 
 \midrule
 DOR & \textbf{0.78}$^\dagger$&\textbf{0.69}$^\dagger$ &0.52 &0.70$^\dagger$&0.66\\
 SSR &\textbf{0.78}$^\dagger$  &\textbf{0.69}$^\dagger$  & \textbf{0.55}$^\dagger$$^\ddagger$& \textbf{0.71}$^\dagger$&\textbf{0.68}\\ 
 \midrule
 LDA & 0.61&0.55 &0.52 &0.64&0.58\\
 LSA & \textbf{0.78}&\textbf{0.69} &0.53 &0.70&0.67\\
 \midrule
 \midrule
 \cite{maharjan2014simple}&0.57&0.66&0.53&0.66&0.60\\
 \cite{villena2014daedalus}&0.64&0.68&0.54&0.51&0.59\\
 \cite{weren2014examining}&0.82&0.71&0.57&0.78&0.72\\
 \bottomrule
 \end{tabular}
 \label{tab:AccuracyGenderVsTopicBased}
 \end{table}





\subsubsection{Comparison against other approaches}

This section presents a comparison of the proposed framework, employing the DOR and SSR distributional term representations, against the works from PAN@2014 which reported results in the training partition. We mainly consider the following three works: \cite{maharjan2014simple}, based in a combination of term n-grams with different $n$ values using the MapReduce programming paradigm; \cite{villena2014daedalus}, which used a two-level classifier composed of a document-oriented classifier with a term vector model representation in combination with a voting strategy; \cite{weren2014exploring}, which considered a method based on information retrieval ideas. We exclude from this comparison our work \cite{pastor14}, the winning approach at PAN@2014 since it was based on the SSR representation.

The bottom rows from Tables \ref{tab:AccuracyAgeVsTopicBased} and \ref{tab:AccuracyGenderVsTopicBased} show the results for the age and gender classification tasks. As it is possible to observe, the employed DTRs outperform the results from \cite{maharjan2014simple} and \cite{villena2014daedalus} in both tasks and for all genres. Nevertheless, they could not improve the results from \cite{weren2014exploring}. It is important to consider that in \cite{weren2014exploring} authors reported the best results obtained after an exhaustive tunning stage, i.e., the selection of the best classification method from a broad family of algorithms, as well as the selection of the best subset of features for each social media genre. Also, this approach obtained an erratic performance during the test phase of PAN@1014 \cite{rangel2014overview}, especially for the Twitter domain, where it achieved an accuracy 15-points lower than our SSR-based approach \cite{pastor14}. Hence, the overall outlook seems to indicate that the proposed framework is more robust than most previous approaches for AP, as some DTRs are nonparametric and therefore they do not require for a tunning stage.

\subsection{Getting to know the learned concepts: a qualitative analysis}

Previous experiments indicate that the DOR distributional representation has several advantages in comparison to other approaches, for example, it does not require tunning any parameter, it allows building relatively compact non-sparse representations, and it obtains very competitive results. Moving a step forward, we performed an analysis of the \textit{interpretability} of DOR. As explained in Section \ref{subsec:dor}, in the DOR representation each document is represented by its relation to other documents. Thus, in the context of AP, it means that each user is represented by its relation to or similarity with other users from the corpus. Accordingly, the features with greater IG are the more discriminative users among the classes (i.e., target profiles).

To exemplify this, Table \ref{tab:ImportantWordsForBlogsGender} shows the top ten words from the three most representative male and female profiles. Words were selected according to their TF-IDF values. As it is shown, each one of these users tends to write about different topics, nevertheless, all they show interesting and important content aspects of their classes. For example, Male 1 writes about books and pictures, Male 2 writes about technology, and Male 3 writes about exercises and diets. In the case of females, notice that Female 1 writes about networks, accessories, and shopping, Female 2 writes about food and drinks ingredients, and Female 3 write about social media management. 


\begin{table*}
 \centering
 \scriptsize
 \caption{The three most representative \textit{male} and \textit{female} users from the blogs genre (used as features in the DOR representation). Showed words correspond to the top ten words according to their TF-IDF value for each user.}
 \begin{tabular}{|c|c|c||c|c|c|}
 \toprule 
 Male 1 & Male 2 & Male 3 & Female 1 &Female 2 & Female 3 \\ 
 \midrule
photo&google&game&bloglovin'&smooth&knowledge\\
book&width&exercise&pinterest&acidity&media\\
draw&sms&breakfast&style&palate&management\\
sketches&windows&baseball&instagram&alcohol&social\\
teaching&smile&salad&twitter&cherry&culture\\
learn&keyboard&dinner&accessories&licorice&change\\
learning&success&eating&facebook&vanilla&content\\
spent&delete&running&necklace&cheese&conversation\\
anxiety&forget&protein&shoes&chocolate&meeting\\
brothers&funny&training&shopping&aromatic&blogs\\
\bottomrule
 \end{tabular}
 \label{tab:ImportantWordsForBlogsGender}
 \end{table*}

\subsection{On the role of the collection characteristics}

In order to enrich the performed analysis, we carried out some initial experiments for analyzing the role or influence of different characteristics from the collections over the performance of the considered DTRs. In particular, we measured the correlation between the value of these characteristics and the improvement in accuracy of the DTRs over the baseline result.

\begin{itemize}
\item Type Token Ratio (\textbf{TTR}). It measures the vocabulary richness of the collection as the ratio of different terms to the total number of terms in the collection \cite{laufer1995vocabulary}. 

\item Lexical Density (\textbf{LD}). This is another vocabulary richness measure. It is calculated as the ratio of content terms (nouns, verbs, adjectives, adverbs) to the total number of terms in the collection \cite{laufer1995vocabulary}.

\item Sophistication (\textbf{SX}). It indicates the percentage of sophisticated terms compared with the number of terms in the collection. A term is considered as sophisticated if its length is greater than the average terms length with a standard deviation \cite{lu2012relationship}; 

\item Shortness (\textbf{S}). It is calculated as the arithmetic mean of the lengths of the document from the collection \cite{tellez2009defining}. Thus, the higher its value, the longer the documents.

\item Class imbalance (\textbf{In}). It is calculated as the standard deviation of the differences between the current and the ideal number of documents for each category. The ideal number of documents per category is defined as the ratio of the number of documents in the collection and the number of categories. The higher the value of class imbalance, the more unbalanced the collection is \cite{tellez2009defining}. 

\item Hardness (\textbf{H}). It measures the vocabulary overlap among the texts from the different categories (profiles) in the collection. A collection is harder to classify if users from different profiles share much vocabulary. As well as if users from the same class write about very different things. For its computation, we considered all the combinations of two categories from the collection, and for each of them, we calculated the text overlap average. The text overlap is calculated using the Jaccard coefficient \cite{pinto2007relative}.

\end{itemize}

\begin{figure*}[t]
  \centering
    \includegraphics[width=5.8in]{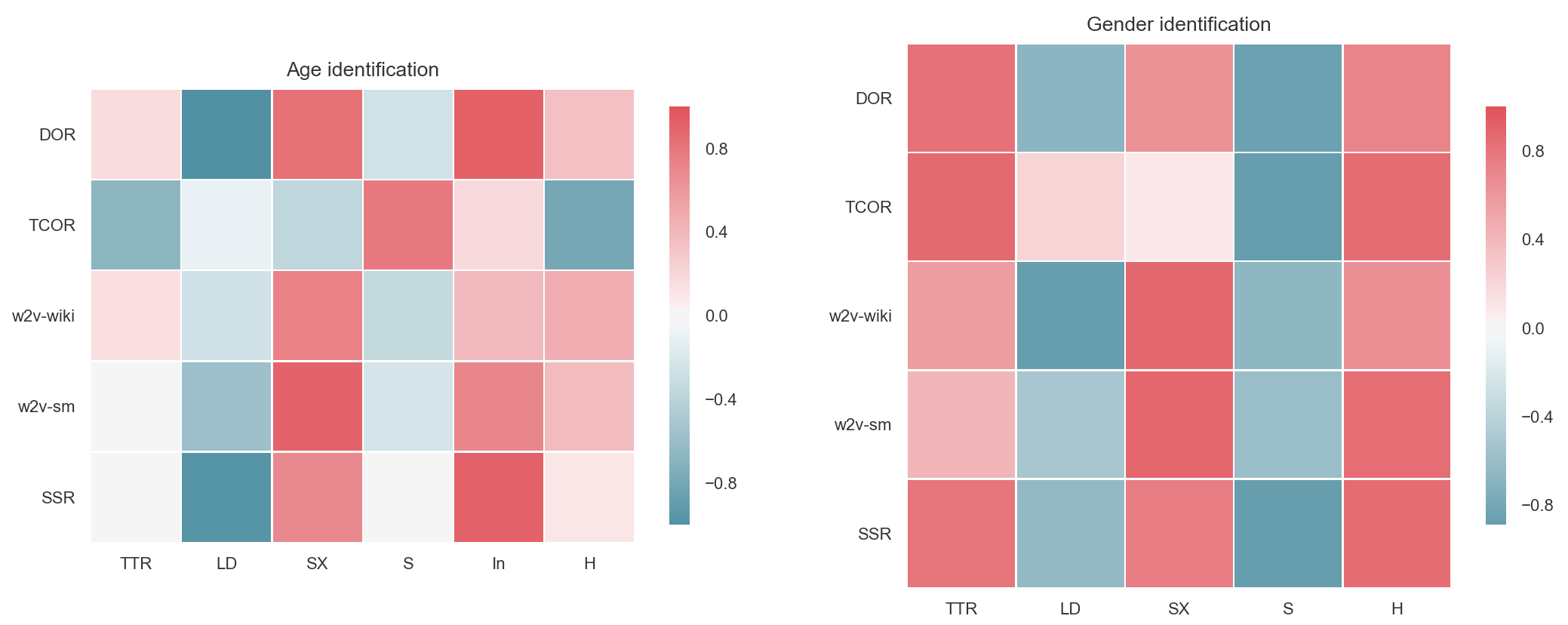}
    \caption{Correlation map between the obtained improvement from all considered DTRs and several collection characteristics. The accuracy improvement is obtained by comparing the result of each DTR against the BoW method.}
  \label{fig:correlationMap}
\end{figure*}

Figure \ref{fig:correlationMap} shows two heat maps that indicate the level of correlation between the evaluated characteristics and the improvement in accuracy of the DTRs over the BoW approach. Although this analysis was basic and straightforward, it helped in discovering that all DTRs, except TCOR, show some common patterns. On the one hand, the figures show a positive correlation concerning the vocabulary sophistication (SX) and the collection hardness (H). That means that the DTRs tend to obtained better results than the BOW for collections having more strange, unusual words as well as for collections showing a considerable overlap among the vocabulary of the different profiles. On the other hand, they show a negative correlation with the shortness (S) and lexical density (LD) characteristics, indicating that DTRs tend to obtained better results than BoW for collections of short texts containing few content terms. Also, this analysis shows that, for the age classification problem, DTRs results positive correlate with the class imbalance (In). All these identified characteristics of the DTRs are important since it is quite common to have imbalanced training sets, short texts and a lot of unusual words in most social media applications.



\section{Conclusions}
\label{sec:Conclusions}

Author profiling consists of analyzing written documents to determine relevant demographic information from their authors. Through the years, the majority of research work has focused on identifying and extracting relevant features for building a suitable representation that learns the textual patterns of distinct profiles. In this same line, a common conclusion among previous research is that content-based features provide more important information than style-based features. 

By previous findings, this work aimed to determine the pertinence of using distributional term representations (DTRs) for solving the author profiling task. Our intuition was that utilizing capturing the semantics of the documents by exploiting the distributional hypothesis, it is possible to identify more discriminative content-based information. Thus, we proposed a novel framework for supervised author profiling in social media domains using DTRs, mainly, we studied for the first time the DOR and TCOR representations and compared their performance against other popular DTRs as well as against two well-known topic-based approaches, namely LDA and LSA.  

For our experiments, we considered the PAN@2014 dataset, which was specially designed for evaluating the AP problem in social media domains. The obtained results indicate that DTRs are suitable for the AP task in social media domains. Mainly, the DOR representation achieved the best accuracy results, while showing the best level of interpretability, since more discriminative features are representative authors from each class. Moreover, a detailed analysis of these results shows that they are robust to class imbalance as well as able to take advantage of the different characteristics of social media data such as their shortness and lexical richness.

A significant advantage of the proposed framework using DTRs is its robustness across different social media genres. Contrary to LDA and LSA, our proposed approach using the DOR term representation does not require any tunning phase, which is a tremendous benefit for social media applications. 

Finally, this work represents the first attempt for carefully determining the pertinence of distinct DTRs approaches for the AP task as well as for analyzing the impact of topic-based methods by its own, which has not been done before. In concordance with previous research, our obtained results allow us to assert that content-based features are  attributes for the AP task. The performed analysis in this work will help future research in AP since it represents a thoughtful and broad study on popular forms of representations. 







\section*{Acknowledgements}

This work was partially supported by CONACYT under projects grant CB-2015-01-258588, CB-2015-01-257383 and FC-2410. Also \'Alvarez-Carmona thanks for doctoral scholarship CONACyT-Mexico 401887.


\bibliographystyle{plain}
\bibliography{references}

\end{document}